\documentclass{ifacconf}

\usepackage{fancyhdr}
\newcommand{\mytitle}{\textbf{Accepted final version.}
To appear in \textit{Proceedings of the IFAC World Congress, 2026}.\\
\copyright 2026 the authors. This work has been accepted to IFAC for publication under a Creative Commons Licence CC-BY-NC-ND}
\usepackage{graphicx}      
\usepackage{natbib}        

\usepackage{subcaption}
\usepackage{booktabs}
\usepackage{amsmath}
\usepackage{mathrsfs}
\usepackage{url}

\usepackage{longtable}

\usepackage{amsfonts}

\usepackage{amssymb}
\newtheorem{lemma}{Lemma}[section]
\newtheorem{definition}{Definition}[section]
\usepackage{multirow}
\usepackage{xcolor}
\usepackage{colortbl}
\usepackage{placeins}
\usepackage{float}
\usepackage[ruled,vlined,algo2e]{algorithm2e}
\begin{document}
\begin{frontmatter}

\title{Individual Control Barrier Functions-Guided Diffusion Model for Safe Offline Multi-Agent Reinforcement Learning} 


\author[First]{Qingyun Guo} 
\author[First]{Junyi Shi} 
\author[Second]{Jianuo Huang}
\author[Third]{Tianyu Shi}

\address[First]{Department of Electrical Engineering and Automation, Aalto University, Espoo, Finland (e-mail: firstname.lastname@aalto.fi)}
\address[Second]{School of Computing and Data Science, Xiamen University Malaysia, Sepang, Malaysia}
\address[Third]{Department of Computer Science, University of Toronto, Toronto, Canada (email: firstname.lastname@mail.utoronto.ca)}

\begin{abstract}
Offline reinforcement learning allows control policies to be learned directly from data without online interaction, making it suitable for safety-critical tasks. Recent studies have applied diffusion models to offline reinforcement learning to leverage their strong capacity for modeling complex data distributions. However, existing approaches primarily focus on single-agent settings, leaving the safety challenges in multi-agent environments largely unexplored. In this work, we propose a safe offline multi-agent reinforcement learning algorithm that embeds neural individual control barrier functions into the diffusion model to enhance safety during trajectory generation, with control policies recovered through inverse dynamics. We evaluate our algorithm across diverse benchmarks, demonstrating substantial safety improvements while maintaining competitive rewards.
\end{abstract}


\end{frontmatter}
\thispagestyle{fancy}
\pagestyle{empty}
\section{INTRODUCTION}

Autonomous systems must learn control policies that meet strict safety requirements in safety-critical applications such as robotics~\citep{act13090364,thomas2022safereinforcementlearningimagining, shi2025event}, autonomous driving~\citep{ELALLID20227366,isele2019safereinforcementlearningautonomous}, and vehicle scheduling~\citep{BASSO2022102496}. Safe reinforcement learning (RL) is a promising approach for solving decision-making problems in autonomous systems under safety constraints. Various frameworks exist in safe RL, with a common approach formulating the problem as a Constrained Markov Decision Process (CMDP)~\citep{altman-constrainedMDP}, where a cost function quantifies safety violations, restricting policy search to a feasible safe region~\citep{wang2023enforcinghardconstraintssoft}. Other methods incorporate control theory, leveraging Lyapunov functions~\citep{Zhang2020LyapunovBasedRL} or Control Barrier Functions (CBFs)~\citep{10161418} to guide policy learning.

However, most of the research focuses on safe online RL in single-agent~\citep{achiam2017constrainedpolicyoptimization} or multi-agent~\citep{qin2021learningsafemultiagentcontrol,berducci2023learningadaptivesafetymultiagent} settings. Although online safe RL methods have shown promising results, including in real-world robotic control \citep{10530650}, they still require interaction with the environment during learning~\citep{zheng2024safeofflinereinforcementlearning}, which can be costly and risky in safety-critical applications. As a result, real-world safety-critical applications remain hesitant to adopt safe online RL solutions.

A promising alternative is safe offline RL, where policies are learned from pre-collected datasets without any online interaction. Recently, generative models, diffusion models in particular, have become popular in safe offline RL for their ability to capture complex data distributions while mitigating extrapolation errors~\citep{zhu2024madiffofflinemultiagentlearning}. FISOR \citep{zheng2024safeofflinereinforcementlearning} enforces safety constraints by identifying the largest feasible region and generating actions through a diffusion model. SafeDiffuser \citep{xiao2023safediffusersafeplanningdiffusion} incorporates forward invariance into diffusion models to improve safety in motion planning.
Despite these developments, both approaches have limitations. FISOR \citep{zheng2024safeofflinereinforcementlearning} defines the feasible region by learning a cost function, which is hard to estimate in offline settings due to the extrapolation error. SafeDiffuser \citep{xiao2023safediffusersafeplanningdiffusion}, on the other hand, shows results
only in relatively low-dimensional environments. Moreover, both methods are restricted to single-agent settings, and without modeling safety and
coordination among agents, their applicability to multi-agent systems is unclear \citep{zhu2024madiffofflinemultiagentlearning}.

\begin{figure}[tbp]
    \centering
    \includegraphics[width=\linewidth]{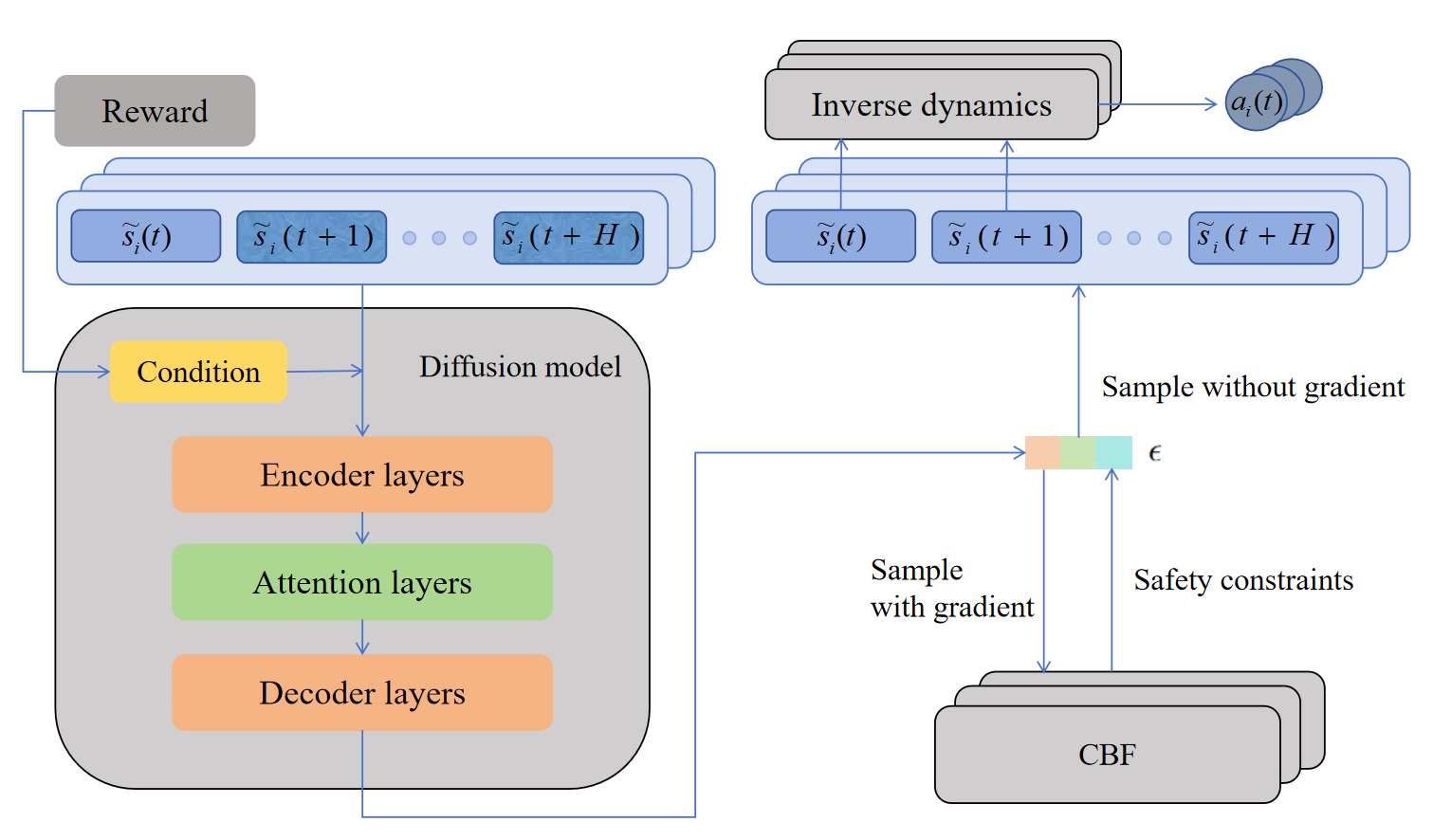}
    \caption{Overview of the proposed framework. Multi-Agent Diffusion Model (MADIFF)~\citep{zhu2024madiffofflinemultiagentlearning} generates reward-conditioned state trajectories, which are guided by individual CBFs. The losses of neural individual CBFs are backpropagated through the diffusion model to guide trajectory generation toward safer, high-reward regions. Policies are then recovered by inverse dynamics.}
    \label{fig:sys}
\end{figure}

To address the dependence on the cost function and the lack of consideration for multi-agent systems, we propose a safe offline multi-agent RL (MARL) algorithm that integrates neural individual CBFs with a multi-agent diffusion model. Our diffusion model follows the centralized training and decentralized execution (CTDE) paradigm \citep{zhu2024madiffofflinemultiagentlearning} to model the coordination among agents. This MARL formulation is important because high-reward behavior in multi-agent systems depends on coordinated interactions among agents, while individual CBFs provide safety guidance. Our algorithm diffuses over states instead of actions, which are easier to model \citep{ajay2023conditionalgenerativemodelingneed}, and is conditioned on rewards to generate high-reward trajectories. Instead of learning a cost function, our algorithm employs neural individual CBFs to guide each agent toward satisfying safety constraints during the training of the diffusion model. In such a way, our algorithm allows the diffusion model to generate high-reward states in safe sets defined by CBFs. The final control policies are inferred through inverse dynamics modules. An overview of the framework is shown in Fig.~\ref{fig:sys}.

Our main contributions are as follows:  
\begin{enumerate}
    \item \textbf{Individual CBFs-Guided Diffusion Model:} We design a CBFs-guided algorithm for the reward-conditioned multi-agent diffusion, in which losses of neural individual CBFs are backpropagated through generated state trajectories. This allows safety constraints to shape the learned trajectory distribution during offline training.  
    \item \textbf{Comprehensive Evaluation:} We conduct extensive comparisons against offline MARL baselines, demonstrating our method’s strong performance in both rewards and safety.  
\end{enumerate}

\section{Preliminaries}
In this section, we introduce preliminaries for our method, including the offline MARL setting, the CBFs for safety constraints, and the diffusion model used for trajectory generation.
\subsection{Offline Multi-Agent Reinforcement Learning}  

The offline MARL problem is formulated as a decentralized partially observable Markov decision process (Dec-POMDP), defined by \(\{N, \mathcal{S}, O, \mathcal{A},\allowbreak f, \rho^0,\allowbreak \gamma, R\}\), where 
\(N=\{1,\ldots,n\}\) is the agent set, \(\mathcal{S}\) is the state space, \(O=\prod_i O_i\) and \(\mathcal{A}=\prod_i \mathcal{A}_i\) are the joint observation and action spaces, 
\(f\) is the transition function, \(\rho^0\) is the initial state distribution, 
\(\gamma\in(0,1)\) is the discount factor, and \(R\) is the shared reward function~\citep{zhu2024madiffofflinemultiagentlearning}. At each time step \(t\), agent \(i\) receives an observation \(o_i(t) \in O_i\) and selects an action \(a_i(t) \in \mathcal{A}_i\) according to its policy \(\pi_i\), the joint action \(\mathbf{a}(t) = (a_1(t), \dots, a_n(t))\) determines the next state \(\mathbf{s}(t+1)\) via \(f\), and agents receive a shared reward \(\mathbf{r}(\mathbf{s}(t), \mathbf{a}(t))\). The objective of this work is to learn policies maximizing
\(
\mathbb{E}\!\left[\sum_{t=0}^{\infty}\gamma^t \mathbf{r}(\mathbf{s}(t),\mathbf{a}(t))\right]\), while avoiding dangerous sets. 
In offline MARL, agents learn from a pre-collected dataset $\mathcal{D}$ containing trajectories $(\mathbf{s},\mathbf{o},\mathbf{a},\mathbf{r})$, which are generated by unknown behavior policies. 
Following \citet{zhu2024madiffofflinemultiagentlearning}, we use bold symbols for joint vectors of all agents.

\subsection{Control Barrier Functions}  

We consider a nonlinear control system with \( N \) agents, where the discrete-time dynamics of each agent \( i \) are given by:  
\begin{equation}
\label{lab:sys}
s_i(t+1) = f_i(s_i(t), a_i(t)),
\end{equation}  
where \( s_i(t) \in \mathcal{S}_i \subseteq \mathbb{R}^{n} \) represents the state of agent \( i \) at time step \( t \), \( a_i(t) \in A_i \subseteq \mathbb{R}^m \) is the control action, and \( t \) denotes the discrete time step. We define the safe set for agent $i$, $\mathcal{S}_{i,s}$, by the superlevel set of $h$:
\begin{equation}
\label{eq:C}
    \mathcal{S}_{i,s} = \{\, s_i \in \mathcal{S}_i \mid h_i(s_i) \ge 0 \,\}.
\end{equation}

\begin{definition}
\label{def:def1}
 For agent \( i \), the set \( \mathcal{S}_{i,s} \) is forward invariant for system \eqref{lab:sys} if, for some \( a_i \in A_i \), any trajectory starting at \( s_i(0) \in \mathcal{S}_{i,s} \) satisfies \( s_i(t) \in \mathcal{S}_{i,s} \), \( \forall t \geq 0 \). If \( \mathcal{S}_{i,s} \) is forward invariant, we say that the system \eqref{lab:sys} is safe \citep{berducci2023learningadaptivesafetymultiagent,xiao2023safediffusersafeplanningdiffusion}.
\end{definition}  

\begin{definition}
A function \( h_i: \mathbb{R}^n \to \mathbb{R} \) is a discrete CBF for agent \( i \) if there exists an extended class-\(\mathcal{K}\) function \( \alpha \) such that, for system \eqref{lab:sys}, the following condition holds:  
\begin{equation}
\sup_{a_i \in A_i} h_i(f_i(s_i(t), a_i(t))) \geq \alpha(h_i(s_i(t))),
\end{equation}  
for all \( s_i(t) \in \mathcal{S}_{i,s} \), where \( \mathcal{S}_{i,s} \) is the forward invariant set defined in Definition \ref{def:def1} \citep{berducci2023learningadaptivesafetymultiagent}.
\end{definition}  

Here, \(s_i\) is used as a generic state variable for presenting the standard individual CBF condition. The safety of agent \( i \) can be maintained using the CBF, which ensures that its system states remain within a predefined safe set. Specifically, given the control policy \( \pi_i \) for agent \( i \) and its corresponding CBF \( h_i \), we define:   \( \mathcal{S}_{i,d} \subseteq \mathcal{S}_i \backslash
 \mathcal{S}_{i,s}\) as the dangerous set. As shown in \citet{7040372}, if the following conditions hold:
\begin{equation}
\label{lab:initial condition}
\begin{aligned}
    & (\forall s_i \in \mathcal{S}_i(0), h_i(s_i) \geq 0) \land (\forall s_i \in \mathcal{S}_{i,d}, h_i(s_i) < 0) \land \\
    & (\forall s_i \mid h_i(s_i) \geq 0, h_i(f_i(s_i(t), a_i(t))) \geq \alpha(h_i(s_i(t)))),
\end{aligned}
\end{equation}
where \( a_i = \pi_i(s_i) \), then the safe state set $\mathcal{S}_{i,s}$ is forward invariant, ensuring that the state remains in the safe set for all \( t \geq 0 \). This guarantees that the agent's state never enters the dangerous state set \( \mathcal{S}_{i,d} \) under the policy \( \pi_i \)~\citep{qin2021learningsafemultiagentcontrol}.

\subsection{Diffusion Model}  

Diffusion models generate data by learning a denoising process from a dataset \(D\). The forward process gradually transforms samples from the dataset into Gaussian noise via a Markov chain: \(q(x^{k+1} | x^k) = \mathcal{N}(x^{k+1}; \sqrt{\alpha^k} x^k, (1 - \alpha^k)I)\), where \(\alpha^k\) is a variance scheduling hyperparameter, and the final state after \(K\) steps is \(x^K \sim \mathcal{N}(0, I)\). The reverse process denoises a noisy sample \(x^k\) using a parameterized Gaussian transition: \(p_{\theta}(x^{k-1} | x^k) = \mathcal{N}(x^{k-1} | \mu_{\theta}(x^k, k), \Sigma_{\theta}(x^k, k))\), where \(\mu_{\theta}\) and \(\Sigma_{\theta}\) are learned by the neural networks. Starting from \(x^K \sim \mathcal{N}(0, I)\), the reverse process iteratively refines the sample to recover data from the target distribution. Training minimizes the difference between predicted and actual noise using the loss function~\citep{ho2020denoisingdiffusionprobabilisticmodels}: \(\mathcal{L}(\theta) = \mathbb{E}_{k, x_0, \epsilon} \left[ \| \epsilon - \epsilon_{\theta}(x^k, k) \|^2 \right]\), where \(\epsilon_{\theta}(x^k, k)\) predicts the noise \(\epsilon \sim \mathcal{N}(0, I)\) added during the forward process. 

A popular framework for diffusion models in RL diffuses over state trajectories only \citep{ajay2023conditionalgenerativemodelingneed}: trajectories \(\boldsymbol{\tau} := [\mathbf{s}(t), \mathbf{s}(t+1), \ldots, \mathbf{s}(t+H-1)]\), where \(H\) is the trajectory length, and actions are inferred using an inverse dynamics module: \(\mathbf{a}(t) = I_{\phi}(\mathbf{s}(t), \mathbf{s}(t+1))\). To avoid notation conflicts, we use superscript \(k\) for diffusion steps, subscript \(i\) for agents, and \(t\) for trajectory timestamps, e.g., \(\tau_i^k := (s_i(t), s_i(t+1), \dots, s_i(t+H-1))^k\). To generate targeted synthetic data (e.g., trajectories with high rewards in RL), the diffusion model conditions on state \( \mathbf{s} \) and information \( y \) using classifier-free guidance, which combines conditional noise \(\epsilon_{\theta}(\mathbf{s}^k, y, k)\) and unconditional noise \(\epsilon_{\theta}(\mathbf{s}^k, \varnothing, k)\) to guide sampling via perturbed noise: \(\epsilon_{\theta}(\mathbf{s}^k, k) + \omega (\epsilon_{\theta}(\mathbf{s}^k, y, k) - \epsilon_{\theta}(\mathbf{s}^k,\varnothing, k))\).
\section{Methodology}
In this section, we present our methodology for safe offline MARL by integrating individual CBFs into the multi-agent diffusion model. We first introduce the multi-agent diffusion model in Section~\ref{sec:DM}. In Section~\ref{sec:CBFs}, we describe the incorporation of individual CBFs, and finally, we discuss policy learning in Section~\ref{sec:PL}.

\subsection{Multi-Agent Diffusion Model}
\label{sec:DM}

Our diffusion model builds on MADIFF~\citep{zhu2024madiffofflinemultiagentlearning}, which uses the CTDE paradigm to model coordination in offline MARL. MADIFF provides both centralized and decentralized trajectory-generation schemes. The former improves coordination with global states but may introduce redundant information and overlook agent-specific details, whereas the latter supports agent-specific generation from local observations but may miss global coordination.

To balance the strengths and weaknesses of these models, we leverage feature-pruned agent-specific global states~\citep{yu2022surprisingeffectivenessppocooperative}. The main idea is that while the global state can enhance coordination among agents, relying on it alone may cause agents to overlook individual tasks. To address this, we augment the global state with each agent’s local observation and prune redundant features between them, preserving both cooperation and individual effectiveness without introducing unnecessary complexity. For an agent $i$, this approach tailors global state ${s}(t)$ information to each agent's local observation $o_i(t)$, while filtering out redundant information. We denote this processed state as $\tilde{s}_i(t)$. The input trajectory for agent $i$ is then given by:
\begin{equation}
    \tau_i = \left[\tilde{s}_i(t), \tilde{s}_i(t+1), \dots, \tilde{s}_i(t+H-1) \right].
\end{equation}

Our diffusion model is conditioned on the reward $\mathbf{r}$ to guide the learning process to focus on states with high rewards. The objective is to minimize the discrepancy between the predicted noise and the true noise introduced in the forward diffusion process. The loss function is \citep{zhu2024madiffofflinemultiagentlearning}:
\begin{align}
\label{eq:DFLF}
    \mathcal{L}^D(\theta) := \mathbb{E}_{k, \boldsymbol{\tau} \in \mathcal{D}, \beta \sim \text{Bern}(p)} 
    \Big[ & \|\epsilon - \epsilon_{\theta}(\boldsymbol{\tau}^k, (1-\beta)\mathbf{r} \notag \\
    & + \beta \mathbf{\emptyset}, k) \|^2 \Big].
\end{align}

This loss function entails generating high-reward trajectories with the diffusion model. However, it does not guarantee that the states within these trajectories remain inside the safe set. To improve safety, we incorporate CBFs into the learning process, which we describe in Section~\ref{sec:CBFs}.

\subsection{Control Barrier Functions Learning}
\label{sec:CBFs}
To identify safe sets with high rewards, we propose a method for learning neural individual CBFs and integrating them with the diffusion model. A key challenge in improving global safety in complex multi-agent systems is that focusing solely on overall system safety may overlook individual agents in critical states. For example, an agent near a hazard may not receive timely intervention, potentially entering a danger zone and triggering a chain reaction that affects the entire system’s safety. To mitigate such risks, we improve safety by addressing local hazards, thereby supporting the overall safety of the multi-agent system \citep{elsayedaly2021safemultiagentreinforcementlearning}. Accordingly, our algorithm learns the individual CBF for each agent, allowing safety constraints to be embedded directly into the state generation process. Moreover, learning neural CBFs rather than manually designing them enables the algorithm to adapt to different environments and safety constraints.

Our approach addresses task-dependent physical safety constraints, such as collision avoidance and speed regulation. To embed neural individual CBFs into the diffusion model, we keep the generated states with gradients from the diffusion model, denoted as $\tilde{\mathbf{s}}^\theta$. In such a way, each agent's safety constraints can be incorporated into the diffusion model through backpropagation. For each agent \(i\), the CBF is defined on the processed agent-specific state \(\tilde{s}_i\) introduced in Section \ref{sec:DM}. Since \(\tilde{s}_i\) is obtained from the global state and the local observation of agent \(i\) after feature pruning, it contains the interaction-relevant information needed for safety evaluation, such as relative positions for collision avoidance or velocity-related components for speed regulation. For each agent \( i \), the state space consists of the safe set \( \tilde{\mathcal{S}}_{i,s} \), dangerous set \( \tilde{\mathcal{S}}_{i,d} \), initial conditions \( \tilde{\mathcal{S}}_{i,0} \subseteq \tilde{\mathcal{S}}_{i,s} \) \citep{qin2021learningsafemultiagentcontrol}.

\begin{lemma}
If agent \(i\) satisfies its task-specific safety criterion at time \(t\), then agent \( i \) is safe at \( t \). If all agents are safe at \( t \), then the entire multi-agent system is safe at \( t \)\citep{qin2021learningsafemultiagentcontrol}.
\label{le:safe}
\end{lemma}

For the task-specific safety criteria considered in this work, learning neural individual CBFs allows each agent to adhere to its own safety constraints and act in safe sets. Consequently, the safety of the entire multi-agent system is improved according to Lemma \ref{le:safe}.

Based on Equation \ref{lab:initial condition}, we define the individual CBF conditions as:
\begin{equation}
\begin{aligned}
    & (\forall \tilde{s}_i \in \tilde{\mathcal{S}}_i(0), \ h_i(\tilde{s}_i) \geq 0)  \land (\forall \tilde{s}_i \in \tilde{\mathcal{S}}_{i,d}, \ h_i(\tilde{s}_i) < 0) \\
    & \land (\forall \tilde{s}_i \mid h_i(\tilde{s}_i) \geq 0, \ h_i(\tilde{s}_i(t+1)) \geq \alpha h_i(\tilde{s}_i(t))).
\end{aligned}
\end{equation}

To learn the individual CBF and integrate it into the diffusion model learning process, we propose the following loss function:
\begin{equation}
\begin{aligned}
\label{eq:CBFLF}
    &\mathcal{L}_i^C(\omega_i,\theta) = \sum_{\tilde{s}_i \in \tilde{\mathcal{S}}_{i,0}} \max \left( 0, \delta - h_i^{\omega_i}(\tilde{s}^\theta_i) \right)\\ 
    &+ \sum_{\tilde{s}_i \in\tilde{\mathcal{S}}_{i,d}} \max \left( 0, \delta + h_i^{\omega_i}(\tilde{s}^\theta_i) \right)\\
    &+ \sum_{\tilde{s}_i \in \tilde{\mathcal{S}}_{i,s}} \max \Big( 0, \delta -  h_i^{\omega_i}(\tilde{s}^\theta_i(t+1)) + \alpha h_i^{\omega_i} (\tilde{s}^\theta_i(t)) \Big),
\end{aligned}
\end{equation}
where \( \delta \) is a margin parameter \citep{qin2021learningsafemultiagentcontrol}. \( \tilde{s}^\theta_i(t) \) is sampled from the learned noise following the Denoising Diffusion Probabilistic Model (DDPM) schedule with gradients. By incorporating neural individual CBFs into the computation graph of the diffusion model, the losses of CBFs can be backpropagated through the diffusion model, guiding the generated trajectories toward high-reward and safer regions. CBFs provide differentiable safety surrogates that penalize generated trajectories violating barrier conditions. 

The choice of \( \alpha \) determines the conservatism of the CBF constraints. A simple choice is a positive scalar, but in offline MARL, overly conservative safety constraints may cause the diffusion process to be trapped in local optima. To avoid this, we introduce a time-based relaxation term for \( \alpha \), inspired by \citet{xiao2023safediffusersafeplanningdiffusion}:
\begin{equation}
    \alpha = m-\eta_i^k,
\end{equation}
where \( m \) is a positive scalar, \( \eta_i^k \) is a relaxation weight satisfying \(0\leq \eta_i^k \leq m \), which gradually decreases to \( 0 \) as \( k \rightarrow 0 \).

\subsection{Policy Learning}
\label{sec:PL}

So far, we have introduced the method to generate states with high rewards in safe sets. To learn control policies from these states, we use inverse dynamics to infer policies \citep{ajay2023conditionalgenerativemodelingneed}:
\begin{equation}
\label{eq:PLFL}
\mathcal{L}_i^P(\phi_i) =  \mathbb{E}_{\mathcal{D}} \left[ \| a_i - f_{\phi_i}(\tilde{s}_i(t), \tilde{s}_i(t+1)) \|^2 \right].
\end{equation}
The inverse dynamics module then maps CBFs-guided state trajectories to executable actions, so the safety constraints enter policy learning through the generated trajectory distribution. Finally, our algorithm is shown in Algorithm~\ref{alg:SMAD}.

\begin{algorithm2e}
	\caption{Multi-Agent Diffuser with Individual Control Barrier Functions}
	\label{alg:SMAD}
	\SetAlgoLined
    \DontPrintSemicolon
	\textbf{Input:} Diffusion model $\epsilon_{\theta}$, inverse dynamics $f_{\phi_i}$, control barrier functions $h^{\omega_i}$ \;
	
	\While{step $\leq$ step$_{\max}$}{
        Extract trajectories from datasets \;
		\For{$k = K$ \textbf{to} $1$}{
			Update $\theta$ according to the diffusion model loss function~\eqref{eq:DFLF} \;
            \For{\textbf{each agent} $i$}{
			    Update \(\omega_i\) and \(\theta\) using the control barrier loss function~\eqref{eq:CBFLF} computed on denoised \(\tilde{s}^{\theta}_i\)  \;
                Update $\phi_i$ according to the policy learning loss function~\eqref{eq:PLFL} \;
            }
		}
        \For{\textbf{each agent} $i$}{
			Extract $(\tilde{s}_i(t), \tilde{s}_i(t+1))$ from generated samples \;
            Compute action: $a_i(t) = f_{\phi_i}(\tilde{s}_i(t), \tilde{s}_i(t+1))$ \;
            
		}
	}
\end{algorithm2e}

\section{Experimental Results}
In this section, we present experimental results in datasets from various environments. These environments span both low- and high-dimensional scenarios and involve varying numbers of agents, allowing us to comprehensively evaluate the effectiveness of our algorithm.
\subsection{Environments Description}

We evaluate our algorithm in the following simulation environments: Simple Spread\footnote{\url{https://pettingzoo.farama.org/environments/mpe/simple_spread/}} and Safe Multi-Agent Mujoco (Safe MAMUJOCO)\footnote{\url{https://github.com/PKU-Alignment/safety-gymnasium}}.

\begin{figure}[tbp]
\centering
    \includegraphics[width=0.2\linewidth]{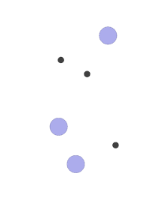}
    \caption{The agents (purple) position themselves to ensure that each one reaches a distinct landmark (black).}
    \label{fig:ss}
\end{figure}

In the Simple Spread environment, as shown in Fig.~\ref{fig:ss}, each agent starts from a random location and aims to cover the landmarks while avoiding collisions. The reward function is the negative of the average distance from each landmark to its nearest agent. If collisions occur, agents receive a shared cost. The dataset for this environment is sourced from \citet{pan2022planbetteramidconservatism} and includes four sets of data collected using policies of varying quality levels, trained by Multi-Agent  Twin Delayed Deep Deterministic Policy Gradient (MATD3)~\citep{ackermann2019reducingoverestimationbiasmultiagent}: Expert, Medium-Replay (md), Medium, and Random.

\begin{figure}[tbp]
\centering
\begin{subfigure}{0.2\textwidth}
   \centering
    \includegraphics[height=3cm]{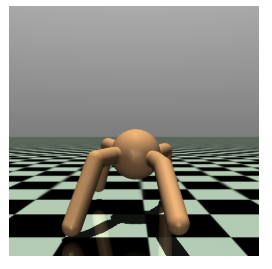}
    \caption{Ant}
    \label{fig:ant}
\end{subfigure}
\begin{subfigure}{0.2\textwidth}
    \centering
    \includegraphics[height=3cm]{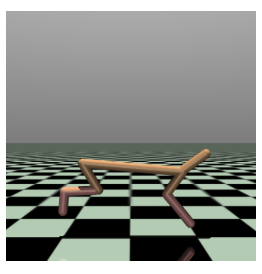}
    \caption{HalfCheetah}
    \label{fig:half}
\end{subfigure}
\caption{(a) Ant: A 3D quadruped robot consisting of a torso with four legs. The goal is to coordinate the legs to move forward. (b) HalfCheetah: A 2D robot with 9 body parts and 8 joints. The objective is to apply torque to the joints to make the cheetah run forward.}
\label{fig:mujoco}
\end{figure}

For Safe MAMUJOCO, the environment factorizes robots shown in Fig. \ref{fig:mujoco} into different subsets, where each part is controlled by an independent agent. The objective is to move the robot forward while maintaining its speed under a predefined velocity threshold. The robot gets a reward based on the distance it moves forward. If the robot's speed exceeds this threshold, agents receive a constant cost. We evaluate our method using three configurations: 2x3 HalfCheetah (2halfcheetah), 2x4 Ant (2ant), and 4x2 Ant (4ant). The datasets are from the Off-the-Grid offline dataset \citep{formanek2023offthegridmarldatasetsbaselines} and include three categories: Good, Medium, and Poor.

\begin{table*}[tbp]
    \centering
    \small
    \renewcommand{\arraystretch}{1.2}
    \setlength{\tabcolsep}{5pt}
    \begin{tabular}{l cc cc cc cc cc cc cc}
        \toprule
        \multirow{2}{*}{Task} & \multicolumn{2}{c}{MA-TD3BC} & \multicolumn{2}{c}{OMAR} &  \multicolumn{2}{c}{Ours}  \\
        & reward $\uparrow$ & cost $\downarrow$ & reward $\uparrow$ & cost $\downarrow$ & reward $\uparrow$ & cost $\downarrow$ \\
        \midrule
        2ant-good & 0.73$\pm$0.11 & 2.39$\pm$0.08 & 0.34$\pm$0.09 & 1.25$\pm$0.07 & \textbf{1.05}$\pm$ 0.10 & \textcolor{blue}{0.44}$\pm$ 0.04  \\
        2ant-medium & 0.41$\pm$0.06 & 1.45$\pm$0.07 & 0.18$\pm$0.06 & \textcolor{blue}{0.35}$\pm$0.05 & \textbf{0.60}$\pm$ 0.09 & 0.80$\pm$ 0.07 \\
        2ant-poor & 0.21$\pm$0.12 & 0.55$\pm$0.10 & 0.20$\pm$0.09 & 0.43$\pm$0.10 & \textbf{0.33}$\pm$ 0.07 & \textcolor{blue}{0.29}$\pm$ 0.05  \\
        4ant-good & 0.70$\pm$0.15 & 1.95$\pm$0.12 & 0.15$\pm$0.10 & \textcolor{blue}{0.25}$\pm$0.07 & \textbf{0.99}$\pm$0.08 & 0.64$\pm$0.06  \\
        4ant-medium & 0.50$\pm$0.12 & 1.73$\pm$0.10  & \textbf{0.69}$\pm$0.15 & 0.82$\pm$0.07 & 0.48$\pm$0.09 & \textcolor{blue}{0.56}$\pm$0.05  \\
        4ant-poor & 0.25$\pm$0.06 & 0.74$\pm$0.05 & 0.23$\pm$0.03 & 0.53$\pm$0.04 & \textbf{0.40}$\pm$0.02 & \textcolor{blue}{0.44}$\pm$0.04  \\
        2halfcheetah-good & \textbf{1.11}$\pm$0.27 & 2.14$\pm$0.12 & 0.81$\pm$0.24 & 1.76$\pm$0.09 & 1.08$\pm$0.16 & \textcolor{blue}{0.66}$\pm$0.07  \\
        2halfcheetah-medium & \textbf{0.51}$\pm$0.03 & 1.58$\pm$0.04 & 0.25$\pm$0.06 & 1.51$\pm$0.02 & 0.43$\pm$0.05 & \textcolor{blue}{0.48}$\pm$0.01  \\
        2halfcheetah-poor & 0.09$\pm$0.01 & 0.20$\pm$0.02 & 0.05$\pm$0.01 & 0.19$\pm$0.02 & \textbf{0.13}$\pm$0.02 & \textcolor{blue}{0.17}$\pm$0.01  \\
        simple spread-expert & 1.08$\pm$0.02 & 3.30$\pm$0.20 & \textbf{1.35}$\pm$0.02 & 5.50$\pm$0.25 & 1.11$\pm$0.04 & \textcolor{blue}{0.00}$\pm$0.00  \\
        simple spread-md & 0.32$\pm$0.04 & 1.30$\pm$0.10 & 0.60$\pm$0.06 & 2.30$\pm$0.13 & \textbf{0.82}$\pm$0.02 & \textcolor{blue}{0.30}$\pm$0.01 \\
        simple spread-medium & 0.71$\pm$0.05 & 2.10 $\pm$0.13& 0.65$\pm$0.07 & 2.50$\pm$0.15 & \textbf{0.88}$\pm$0.05 & \textcolor{blue}{0.00}$\pm$0.00  \\
        simple spread-random & 0.23$\pm$0.01 & \textcolor{blue}{0.00}$\pm$0.00 & 0.49$\pm$0.03 & 1.50$\pm$0.04 & \textbf{0.56}$\pm$0.01 & \textcolor{blue}{0.00}$\pm$0.00  \\
        \bottomrule
    \end{tabular}
    \caption{Performance comparison across different algorithms on various benchmarks. For each entry, the first number represents the mean over 5 seeds, and the second number indicates the corresponding standard deviation. Higher reward ($\uparrow$) and lower cost ($\downarrow$) are better. The best reward per task is bolded, and the best cost is highlighted in blue.}
    \label{tab:results}
\end{table*}

\subsection{Baselines and Metrics}  
To evaluate the performance of our algorithm, we use the episode reward and cost as metrics, obtained by deploying our learned policy in the online environment. A higher episode reward means better performance, while a lower cost indicates improved safety. Our diffusion model is built upon MADIFF \citep{zhu2024madiffofflinemultiagentlearning}. Each agent's CBF is parameterized by a three-layer fully connected network with 128 hidden units per layer and ReLU activation function. We normalize both cost and reward following \citet{zheng2024safeofflinereinforcementlearning}. The normalized cost is computed as: $c_{\mathrm{normalization}} = \frac{c_{\pi} + k}{l + k}$,
where \( l \) is a human-defined cost limit, and \( k \) is a positive constant to ensure numerical stability when \( l = 0 \). The normalized reward is given by: $
r_{\text{normalization}} = \frac{r - r_{\min}}{r_{\max} - r_{\min}}$. In the same task, we use the same values of \( l \), \( k \), \( r_{\min} \), and \( r_{\max} \) for all methods. Given the lack of safe offline MARL baselines, we compare our approach to state-of-the-art offline MARL algorithms, including OMAR \citep{pan2022planbetteramidconservatism}, and MA-TD3BC \citep{fujimoto2021minimalistapproachofflinereinforcement}. We evaluate all methods with 5 different random seeds.

\subsection{Results}

The evaluation results are presented in Table \ref{tab:results}. Our algorithm achieves the lowest costs in almost all tasks while maintaining competitive rewards. In high-dimensional environments with datasets of varying quality, such as 2ant, our algorithm attains the highest rewards while keeping safety costs low, showing its ability to avoid dangerous regions and learn effective control policies. In the low-dimensional Simple Spread environment, it achieves the best performance on the medium-replay, medium, and random datasets with almost no collisions, demonstrating the effectiveness of neural individual CBFs. In the 4ant and 2halfcheetah tasks, our method also maintains competitive performance.

When dataset quality is high, MA-TD3BC \citep{fujimoto2021minimalistapproachofflinereinforcement} attains high rewards because it relies heavily on supervised learning through behavior cloning. However, its performance degrades substantially as the dataset quality worsens. For instance, in the Simple Spread–random task, MA-TD3BC \citep{fujimoto2021minimalistapproachofflinereinforcement} achieves the lowest cost but obtains only about half the reward of other methods, suggesting that agents may fail to reach the landmarks and thus avoid collisions by not completing the task. OMAR \citep{pan2022planbetteramidconservatism}, on the other hand, tends to fall into local optima due to its local regularization. Although it achieves the lowest cost in the 4ant-good and 2ant-medium settings, it also gets the lowest reward among all algorithms. In contrast, our algorithm consistently achieves a high reward while keeping the cost low across nearly all tasks.

\section{Conclusion}
In this paper, we propose a novel safe offline MARL algorithm that embeds neural individual CBFs into the diffusion model. By learning neural individual CBFs, our approach introduces safety constraints inspired by CBFs without relying on learned cost functions, thereby reducing the impact of extrapolation errors commonly seen in offline RL and improving adaptability across different safety constraints. Nevertheless, the reward and CBF guidance may still bias generated trajectories and lead to out-of-distribution issues. With the diffusion model, our algorithm shows strong performance in various benchmarks. Due to the lack of established benchmarks for safe offline MARL, our evaluation is limited to a set of representative environments. In future work, we will focus on extending our algorithm toward certified closed-loop safety, for example, by combining it with verification modules under sensing uncertainty, model mismatch, and asynchronous multi-agent interactions.

\begin{ack}
We acknowledge the computational resources provided by the Aalto Science-IT project.
\end{ack}
\bibliography{ifacconf}            
\end{document}